\documentclass[twocolumn,10pt,letterpaper,notitlepage]{article}
\usepackage{todonotes}
\usepackage[margin=1in]{geometry}
\usepackage{graphicx}
\usepackage[table]{xcolor}
\usepackage[bigtitle,smallsection,SECTION,absheading]{ahsmeeting}
\usepackage{wrapfig}
\usepackage[colorlinks=false]{hyperref}
\usepackage{xspace}
\usepackage{comment}
\usepackage{enumitem}
\usepackage{subcaption}
\usepackage{mathtools}
\usepackage{amsmath}
\usepackage{amssymb}
\usepackage{mathtools}
\usepackage{tikz}
\usepackage{ragged2e}
\usepackage{pgfplots}
\usetikzlibrary{positioning,quotes,arrows.meta,bending,shapes.misc,shapes.geometric,positioning, calc, backgrounds, fit}

\definecolor{halfgreen}{RGB}{0,128,0}
\definecolor{ahsred}{RGB}{192,0,0}

\newcounter{isorefi}
{\endlist}

\newcommand{\arp}{{ED-324}\xspace}

\newcommand{\dosw}{{DO-178C/ED-12C}\xspace}
\newcommand{\dotool}{{DO-330/ED-215}\xspace}
\newcommand{\ucf}[1]{\texttt{#1}}
\newcommand{\usecase}[4]{\ucf{#1}-\ucf{#2}-\ucf{#3}-\ucf{#4}}

\newcommand{\weight}{{H/C weight estimator}\xspace}

\newtheorem{definition}{Definition}
\newtheorem{objective}{Objective}

\begin{document}

\title{Towards On-Board Implementation of ML-Based Helicopter Weight Estimator}

\author{\textbf{Nicolas Valot} \\PhD student\\
Airbus Helicopters \\ Marignane, France
\and
\textbf{Ammar Mechouche}  \\Analytics and Big Data Expert \\
Airbus Helicopters \\ Marignane, France
\and
\textbf{Benjamin Lesage} \\Researcher \\
ONERA \\Toulouse, France
\and
\textbf{Claire Pagetti} \\Researcher \\
ONERA \\Toulouse, France
\and
\textbf{Louis Fabre}  \\Senior Expert - CVE\\
Airbus Helicopters \\Marignane, France}

\date{}
\abstract{
This paper focuses on the implementation of a novel supervised Machine Learning model for estimating helicopter weight during takeoff, utilizing extensive datasets from Airbus’s global in-service fleet.
The study details a learning assurance process aligned with the EASA concept paper for machine learning application, and with the on-going Eurocae ED-324.
We propose a set of Machine Learning Requirements, a Machine Learning Model Description, and its implementation
for a long short-term memory recurrent neural network. Finally, we verify the requirements on the implementation.
%both addressing safety assessment objectives and means of compliance for Machine Learning applications.
Demonstrated on legacy avionics computers, the implementation is suitable for the deployment of the developed Machine Learning Model weight estimator on airborne targets for critical functions such as on-board alerting.
}
%optional; not included to conform to the MSWord version
%\meeting{Submitted for the AHS 73rd Annual Forum, Virginia Beach, Virginia,
%USA, May 5--7, 2015.}
\nomeeting

\maketitle
\section{Notation}
\begin{tabular}{ ll }
BSP & Board Support Package\\
CBM & Condition Based Maintenance\\
CI & Confidence Interval\\
CPU & Core Processing Unit\\
DPR & Data Processing Requirements\\
H/C & Helicopter\\
HLR & High Level Requirement\\
HLT & High Level Test\\
LCD & Life-Cycle Data\\
LLR & Low Level Requirement\\
LLT & Low Level Test\\
LSTM & Long term Short Term Memory\\
MAE & Mean Average Error\\
ML & Machine Learning\\
MLC & ML Constituent\\
MLDL & ML Development Life-cycle\\
MLCR & MLC Requirements \\
MLM & ML Model\\
MLMD & MLM Description \\
MLMID & MLM Item Description \\
MTOW & Maximum TakeOff Weight\\
ODD & Operational Design Domain\\
OEW & Operational Empty Weight\\
ONNX & Open Neural Network Exchange\\
PDI & Parameter Data Item\\
PSSA & Preliminary System Safety Assessment\\
RNN & Recurrent Neural Network\\
\end{tabular}

\section{Introduction}
\begin{figure}[h!]
  \centering
%    \resizebox{1\linewidth}{!}{%
  \includegraphics[width=0.50\textwidth, trim= 1cm 0cm 1cm 0cm]{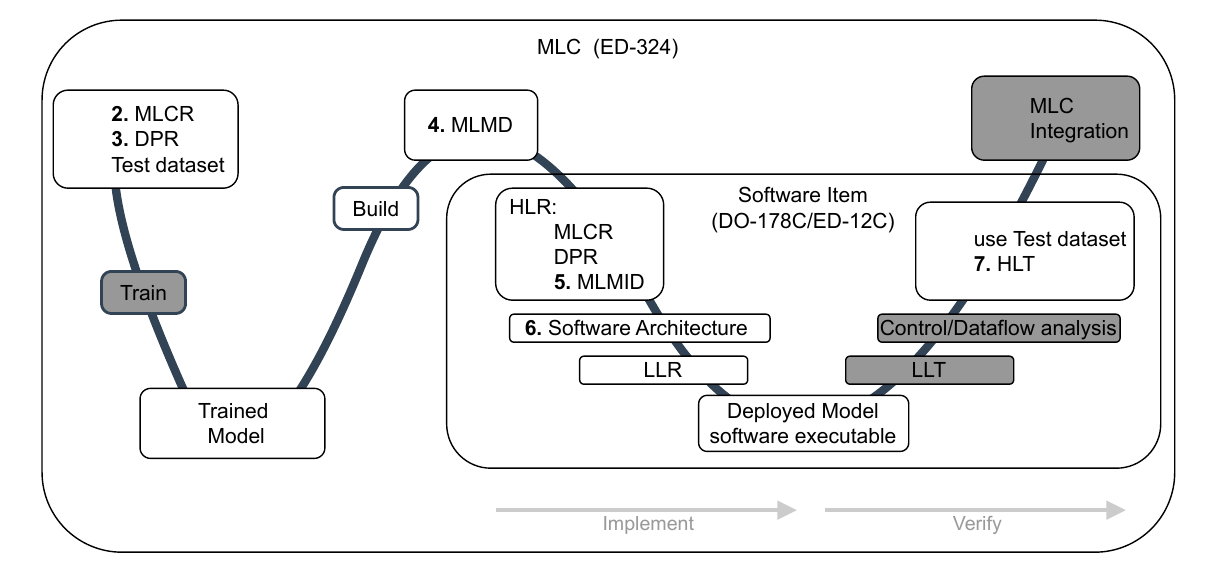}
%  }
	\caption{W-shape development life-cycle, focus on 2nd-V, \dosw Life-Cycle Data (LCD) of the MLC, with its collected \arp LCD as input, the \emph{implementation} downward to the \emph{Deployed Model} software executable and upward to the \emph{verification} \label{fig:do178}}
\end{figure}

This paper presents the implementation of a Machine Learning (ML) helicopter (H/C) weight estimator model through the W-shape development process described in the EASA concept paper~\cite{Easaconcept} and \arp \cite{ARP6983}, as illustrated in Figure~\ref{fig:do178}.
The helicopter (H/C) weight is a key parameter for enhancing safety through advanced Helicopter Terrain Awareness and Warning Systems (HTAWS) and reducing operational costs, particularly within Condition-Based Maintenance (CBM) frameworks. For instance, accurate weight prediction allows for the calculation of components' remaining useful life (RUL) and optimizes the retirement time of mechanical parts.
\begin{figure}[h!]
  \centering
    \resizebox{1.1\linewidth}{!}{%
  \includegraphics{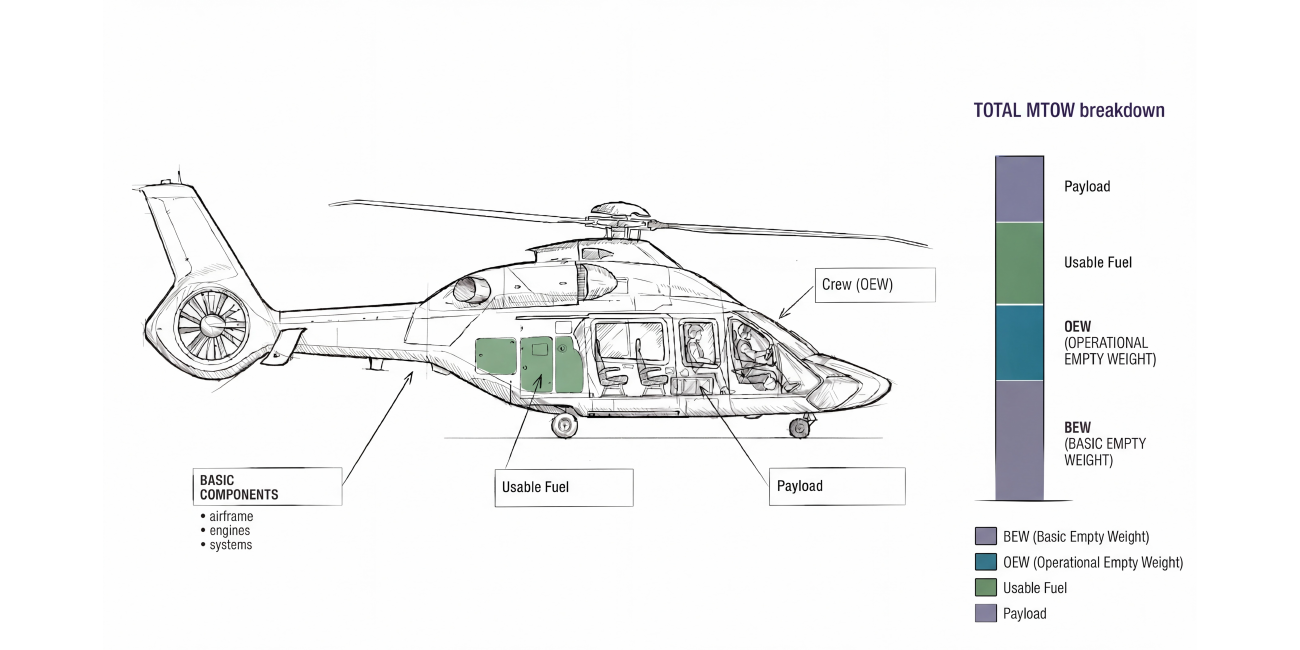}
  }
	\caption{H/C weight breakdown\label{fig:hc_weight}}
\end{figure}

The H/C is certified for a H/C weight being lower than the Maximum TakeOff Weight (MTOW) to meet airworthiness requirements.
As it is done now, there is no on-board sensor
and the aircraft's weight is
estimated with a highly conservative and pessimistic value
that consists of, as illustrated by Figure \ref{fig:hc_weight},
a sum of weight contributions:
\begin{itemize}[before*={\mbox{}\vspace{-.75\baselineskip}}]
  \setlength{\itemsep}{0pt}
    \item the Operational Empty Weight (OEW) composed of:
      \begin{itemize}[before*={\mbox{}\vspace{-.75\baselineskip}}]
        \setlength{\itemsep}{0pt}
      \item the Basic Empty Weight (BEW) defined by the airframe and engine weight plus lubrication and hydraulic fluids, the unusable fuel, and optional equipment weights,
      \item the crew weight,
      \end{itemize}
    \item the fuel (trip and taxi),
    \item the payload (passengers, luggage, cargo, hook load).
\end{itemize}

Machine Learning (ML) may offer new capabilities in airborne systems, and in particular for a \weight.
However, as with any piece of airborne systems,
the safe operation of ML-based systems will have to be guaranteed.
Thus, their development will have to be demonstrated to be compliant with the adequate guidance \cite{Easaconcept, ARP6983}.
Both documents define high-level objectives to confirm
that the ML model (MLM) achieves its intended function,
and that it maintains its training performance in the target environment in a manner commensurate with its allocated safety level.
We assume that the Preliminary System Safety Assessment (PSSA) allocates an item Design Assurance Level C to this \weight ML model.

The Figure \ref{fig:do178} describes the \arp and \dosw life-cycle data. 
The ML Development Life-cycle (MLDL) is the set of processes and activities linked to the MLC design assurance.
The Machine Learning Constituent (MLC) is the entity encompassing the ML model and its Data Processing necessary for the ML model execution.
The MLC requirements (MLCR) are composed of the Operational Domain Description (ODD), and the performance, stability and the robustness requirements.
The Machine Learning Model Description (MLMD) is the output of the ML Model design process composed of the ML model architecture and the model trained parameters.

As a reference, the paper \cite{weightestim} introduces the \weight  use case, and addresses the 1st-V in the \arp W-shape, i.e. the data management, the Operational Design Domain (ODD), the model architecture, its primary design and reports the model's best performances in the model training environment from a data scientist standpoint.

\textbf{Contributions}:
The paper \cite{weightestim} has focused on the EASA concept paper only and as such, has not detailed the requirements of the system and its associated ML constituent as needed by the \arp.
We want to be compliant with the \arp 2nd-V and the implementation, in particular the verification of the implementation, necessitates the ML constituent requirements. 
This is the reason why we first need to complete the 1st-V by defining those requirements.
We then concentrate on the implementation and the interface between the \arp and the \dosw. Such an interface is a tricky question because it is unclear on how to link the HLR (input of the DO) and the requirements of the 1st-V.
Our proposal is to consider them as equal.
Once the HLR are fixed, the implementation is rather classical. The last difficult challenge is how to verify the ML constituent with respect to the HLR (and thus the MLC requirement and the DPR).  Here our strategy relies on a bit-accurate replication of the ML model and statistical verification of ML metric.  
In practice, we:
\begin{itemize}
\setlength{\itemsep}{0pt}
\item specify MLCR and DPR of the 1st-V of the Figure \ref{fig:do178},
\item allocate MLCR and DPR to a software item's \dosw High Level Requirements (HLR) in the 2nd-V,
\item build the MLMD from the trained ML model,
\item decompose the MLMD into a MLM Item Description (MLMID) with a robust Long Term Short Term Memory (LSTM) cell architecture,
\item implement the MLMID with Scade modeling tool and generate C code, 
\item extract model parameters as Parameter Data Item (PDI), which brings the capability of re-training the model without any impact on the software code Life-Cycle Data (LCD). 
\item verify the HLR with High Level Tests (HLT), including the execution time of the ML model on two deployed targets (different processors and compilers).
\end{itemize}

\textbf{Paper Organization}:
The numbers in the Figure \ref{fig:do178} boxes identify their related section in the paper.
We first introduce the \weight in Section \textsl{1.~System Overview}.
We elicit the MLC requirements in the Section \textsl{2.~MLCR} and the Data Processing Requirements in Section \textsl{3.~DPR}.
The Machine Learning Model is described in Section \textsl{4.~MLMD}. 
The MLMD is broken-down in Section \textsl{5.~MLMID}. 
The software architecture is described in \textsl{6.~ML Software Item Architecture}. 
The verification is finally demonstrated in Section \textsl{7.~HLT}, before the conclusion.
The gray boxes of Figure \ref{fig:do178} are legacy \dosw process activities which are not specific to ML and thus not addressed by the paper.

% \newpage
\section{1. System Overview}

  \begin{figure}[h!]
  \centering
    \resizebox{1\linewidth}{!}{%
  \includegraphics{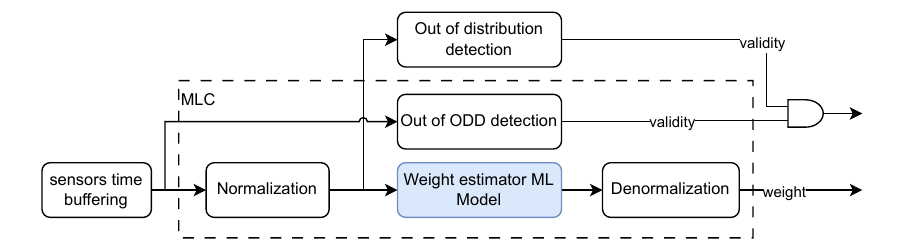}
  }
	\caption{System overview\label{fig:archi}
}
\end{figure}
As there does not exist any airborne weight sensor, the system implementing this function is designed with the Machine Learning Model highlighted in Figure \ref{fig:archi}, as a virtual weight sensor.
The system breaks down into:
\begin{itemize}[before*={\mbox{}\vspace{-.75\baselineskip}}]
\setlength{\itemsep}{0pt}
\item a \emph{MLC} composed of:
\begin{itemize}[before*={\mbox{}\vspace{-.75\baselineskip}}]
\setlength{\itemsep}{0pt}
\item an \emph{Out-of-ODD detection},
\item input data normalization,
\item a \weight ML model,
\item regression output weight denormalization.
\end{itemize}
\item an \emph{Out-of-distribution detection} unsupervised learning model,\footnote{Unsupervised learning is currently out of the \arp scope.}
\end{itemize}

In this paper, we consider only the MLC requirements which specify what is expected from the \weight.

At system level, we allocate some processing resource to the execution of the \weight as a design constraint:

\textbf{MLM-Execution-time}: The MLM (\weight model) shall be computed in less than 1 ms.

The other specified design constraint requirements (e.g. input/output, data recording, in-service monitoring, human machine interface, ...) are out of this paper scope.

\section{2. MLC Requirements}

MLC requirements are necessary to specify \emph{what} is expected from the \weight, and in which boundaries.
The specificity of our \weight is that there is no obvious relationship between input parameters and the weight output, e.g. the weight shall be equal to $\alpha \cdot P_1 + \beta \cdot P_2^2+ \gamma$. 
With the use of a ML model fitting data, it is not possible to \emph{specify an expected output}, given a specific input.
Instead we can \emph{specify statistical output expectations}, e.g. the mean average error between the predictions and the ground truth shall be lower than a threshold.

In the following, we thus first consider the boundaries in which the model operates, the ODD, and the limitations on available datasets within this domain. This supports the definition of requirements on the H/C weight estimator during the operational life of the system.

%In this section, we describe:
%\begin{itemize}
%\setlength{\itemsep}{0pt}
%\item the Operational Design Domain (ODD),
%\begin{itemize}
%\setlength{\itemsep}{0pt}
%\item the ML model input parameters and their  \emph{boundaries},
%\item the datasets representing the ODD population.
%\end{itemize}
%
%\item the \weight requirements:
%\begin{itemize}
%\setlength{\itemsep}{0pt}
%\item the performance requirements are defined with \emph{metrics}, computed out of ML model predictions and the ground truth values of the \emph{Test dataset},
%\item the \emph{robustness} requirements related to the ML model architecture, and the algorihtm \emph{stability} requirement to bound the output variation upon a small input variation.
%\end{itemize}
%\end{itemize}

\subsection{2.1 Operational Design Domain}

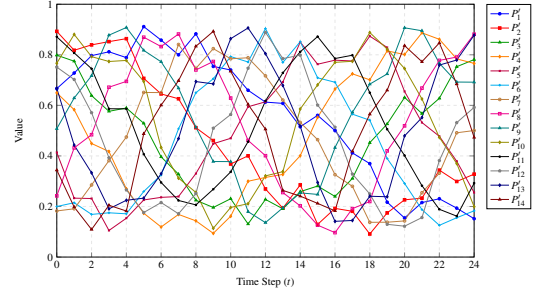
\begin{figure}[htbp]
\centering
\resizebox{0.8\columnwidth}{!}{%
\begin{tikzpicture}
\begin{axis}[
    xlabel={Time Step ($t$)},
    ylabel={Value},
    xmin=0, xmax=24,
    ymin=0, ymax=1.0,
    xtick={0, 2, 4, 6, 8, 10, 12, 14, 16, 18, 20, 22, 24},
    ytick={0, 0.2, 0.4, 0.6, 0.8, 1.0},
    grid=major,
    grid style={dashed, gray!20},
    legend pos=outer north east,
    legend style={cells={anchor=west}},
    width=0.85\textwidth,
    height=0.55\textwidth,
    % Distinct colors and markers
    cycle list={
        {blue, mark=*}, {red, mark=square*}, {green!60!black, mark=triangle*},
        {orange, mark=diamond*}, {purple, mark=x}, {cyan, mark=+},
        {brown, mark=o}, {magenta, mark=square}, {teal, mark=triangle},
        {olive, mark=diamond}, {black, mark=star}, {gray, mark=pentagon*},
        {blue!50!black, mark=diamond}, {red!50!black, mark=triangle}
    }
]

% Generating 14 series: Low-Frequency Wave + Noise
\foreach \i in {1,...,14} {
    \addplot+[
        domain=0:24,
        samples=25,
        line width=0.7pt,
        mark size=1.2pt
    ] 
    % Function breakdown:
    % 0.5: Midpoint
    % 0.35 * sin(...): The low-frequency wave component
    % 0.08 * rand: The noise component (random value between -0.08 and 0.08)
    {0.5 + 0.35 * sin(x * (8 + \i*1.5) + \i*40) + 0.08 * rand};
    \addlegendentryexpanded{$P'_{\i}$}
}

\end{axis}
\end{tikzpicture}
}
\caption{Illustration of one \emph{normalized} sample of ODD: a 25 time steps frame of flight parameter $\{P'_1\ldots P'_{14}\}$ used as an input to \weight ML model.\label{fig:time_series}}
\end{figure}

The ODD of the \weight is characterized by the flight envelope of the H/C, i.e. 14 flight parameters (e.g. speed, altitude, engine power,...)
% denoted $\{P_1,\ldots,P_{14}\} \in \mathbb{R}\times\ldots\times\mathbb{R}$,
which values are recorded by H/C sensors.
% In the avionics,
Each flight parameter sensor is sampled at a frequency specified by system requirements.
The ODD time frame is thus composed of 25 time steps, with a 2Hz sampling frequency,
each composed of the 14 parameters.
We denote $P_{t,i}$ the value of the $i^{th}$ parameter in time step $t$,
and a single time frame is denoted $\{P_1,\ldots,P_{14}\}$, omitting $t$ when there is no ambiguity.
The ODD boundary is thus defined by the boundaries of each parameter $P_{i,t} \in[\underline{P_i}, \overline{P_i}] \forall i\in[1,14], \forall t$.
The \emph{sensor time buffering} in Figure \ref{fig:archi} performs a  resampling for each sensor parameter to fit to the ODD time frame sampling frequency.
An example of \emph{normalized} $\{P'_1,\ldots~P'_{14}\}$ parameter time frame is illustrated in Figure \ref{fig:time_series}.

The ODD boundaries might be narrowed compared to the system input parameter ranges because:
- some input areas are insufficiently covered by datasets,
- ML model behaves poorly on some input areas.
This is the role of the \emph{Out of ODD detection} function in the Figure \ref{fig:archi},
  which is specified by a \emph{derived requirement}:

\textbf{ODD-out-of-domain-detection}:
the MLC shall output \emph{validity=True} when ML model inputs are in ODD, else \emph{validity=False}.
This requirement is defined as a set of rules (e.g. $P_1<\alpha~\text{and}~P_{10}>\beta$) to invalidate model output.

To train the ML model, and verify its performance requirements, some datasets are sampled in the ODD. We call them the ODD population:
\begin{definition}[ODD population]
The ODD population is the set of any possible data in the ODD.
It is composed of:\\
%\begin{itemize}[before*={\mbox{}\vspace{-.75\baselineskip}}]
%\setlength{\itemsep}{0pt}
-~Training data: data which are used to train ML model parameters to fit the ground truth,\\
-~Test data: data which are used to test (validate) model's performance, \textbf{and} were not used to train the ML model,\\
-~Out-of-sample data: data which are neither \emph{seen} during training nor test. Typically, these are data \emph{seen} during the operational life of the system.
%\end{itemize}
\end{definition}
\subsection{2.2 \weight requirements}
\emph{Generalization} in \arp requires the MLC to satisfy its performance requirements on any \emph{out-of-sample} data. We extend this definition to all requirements (e.g. data processing, robustness, stability).

\begin{definition}[Generalization]
The ability of the MLC to satisfy their requirements on ODD population.
\label{def:gene}
\end{definition}

%Before specifying the metrics, we specify the required resolution of the ground truth  in the datasets as a Data Quality Requirement (DQR), which is defined in the first-V of the life-cycle, and is a pre-requisite to the ML model training. It also bounds the best performance which can be expected from the model.

%\textbf{DQR-Ground-truth-resolution\&accuracy}:
%The normalized ground truth resolution and accuracy shall be lower than $5\cdot10^{-3}$.
%We also assume that sensors which provide $P_1\ldots P_{14}$ input parameters satisfy this resolution and accuracy requirement.

We define the metrics which are necessary to elicit requirements:
\begin{definition}[Mean Average Error (MAE) metric]
The MAE measures the mean distance between regression prediction $\hat y$ and the ground truth $y$:
{\small
\begin{equation}
\label{eq:MAE}
\text{MAE} = \frac{\|y - \hat y\|_1}{n} = \frac{1}{n} \sum_{i=1}^n |y_i - \hat y_i|
\end{equation}
}%
\label{def:mae}
\end{definition}

\begin{definition}[Explained Variance ($\text{R}^2$) metric]
The $\text{R}^2$ specifies the explained variance, i.e. the amount of the model output $\hat y$ variance which is associated to the ground truth $y$ variance.
{\small
\begin{equation}
\label{eq:R2}
\text{R}^2 = 1 - \frac{\sum_{i=1}^n\left(y_i-\hat y_i\right)^2}{\sum_{i=1}^n\left(y_i- \overline y\right)^2}
\end{equation}
}%
\label{def:r2}
\end{definition}

Then, we define the \weight ML model:
\begin{definition}[\weight model application]
The neural network named \emph{weight} is a function:\\
\begin{tabular}{llll}
weight:&$\mathbb{R}^{25\times14}$ & $\rightarrow$ & $\mathbb{R}$\\
&$\{P'_{0,1},\ldots,P'_{24,14}\}$ & $\mapsto$ & $\text{weight}(P'_{0,1},\ldots,P'_{24,14})$
\end{tabular}

\label{def:model}
\end{definition}

Then we elicitate MLCR below:

\textbf{MLCR-Performance}:
a performance requirement is specified with a regression metrics threshold.
The regression metrics threshold shall be satisfied by the independent \emph{Test dataset} (a set of data which was not used  during the training), which is provided by the first-V activities.

In the paper \cite{weightestim} table 1,2 and 3, the metric MAE is specified \emph{Normalized Relative to the MTOW}.
We specify the requirement:
{\small
\[
\frac{\text{MAE}}{\text{MTOW}} < 3.10^{-2},\quad\text{R}^2  > 80\%
\]
}%
The \emph{Test dataset} independence is a necessary condition to bring the \emph{generalization} property of Definition \ref{def:gene}, but it is not sufficient.
The \textbf{MLCR-Performance} requirement is based on statistics related to a finite set (\emph{Test dataset}). To \emph{generalize} to the \emph{ODD population}, the number of samples of the \emph{Test dataset} shall be sufficient to ensure that MAE leads in a specified confidence interval, thanks to Central Limit Theorem (CLT).
We propose the following confidence interval requirement:

\textbf{MLCR-oos-confidence-interval}:
The probability P that the \emph{ODD population} metric M leads into an interval of $\pm5\%$ of its value computed for the \emph{Test dataset} $\text{M}^{(T)}$, shall be at least 99\%, i.e.
{\small
\begin{align*}
P\left(0.95\cdot\text{M}^{(T)}<\text{M}<1.05\cdot\text{M}^{(T)}\right) > 0.99\\
\end{align*}
}%
The $\pm5\%$ interval values for our metrics lead to:
{\small
\begin{align*}
\frac{\text{MAE}}{\text{MTOW}} &< (3 + 3*5\%) \cdot10^{-2} = 3.15\cdot10^{-2} \\
\text{R}^2 &> (80 - 80*5\%)  \% = 76 \%
\end{align*}
}%

Stability is a critical factor influencing both model training and inference. While training stability refers to a model's robustness against minor perturbations in the training data, inference stability concerns its resilience to numerical inaccuracies. This paper focuses specifically on the \emph{numerical stability of inference}, examining how rounding error propagation, numerical absorption, and cancellation (all inherent to algorithmic implementation), impact performance.
Stability error is a component of total model error; we aim to ensure its contribution remains negligible.

\textbf{MLCR-Stability}: 
%f(x)-\hat y
We specify the MLC numerical error upper bound requirement:
{\small
\begin{align*}
&\forall P'_{0,1}\ldots P'_{24,14} \in [0, 1]^{25\times14}, \\
&|\text{weight}(P'_{0,1}+\epsilon,\ldots,P'_{24,14}+\epsilon) - \text{weight}(P'_{0,1},\ldots,P'_{24,14})| < 10^{-3}
\end{align*}
$\epsilon$ is the Unit In the Last Place (ULP), i.e. the resolution of  the interval $[0,1]$.
}%

The intermediate computations of the ML model implementation might lead to numerical overflows, occuring in elementary sum of products which are the atoms of neural networks.
We specify the need to avoid this situation.

\textbf{MLCR-Robustness}:
For any input values in the specified range of the parameters $P_1\ldots P_{14}$, the MLC inference shall not exceed MLC layers specified intermediate ranges and output range.

\section{3. MLC Data Processing Requirements}
\label{sec:DPR}

Our \weight parameters $P_1$ to $P_{14}$ represent physical sensor values which are defined with their own ranges, e.g. speed $\in~[0,90~m~\cdot~s^{-1}]$, altitude $\in~[0,6000~m])$. 
Trying to fit data with these ranges would bias the model to learn more from altitude than from speed. 
To eliminate this bias, we normalise the ranges of all parameters.
This is specified by Data Processing Requirements (DPR).
The DPR, attached to the MLC in Figure \ref{fig:do178}, specify the required ML Model's pre and post processing.
The MLC-DPR are composed of 2 categories : i) the DPR required by the data scientist to pre-process data before the learning process. ii) the DPR required in the target system to pre-process the data input from a sensor or from another system and post-process the data output before sending it to the user system.
As i) is handled in the 1st V-Cycle, in this paper, we focus on ii), and more specifically on normalization DPR. The normalization is a bijective application:

\begin{tabular}{rcl}
 $\text{denormalize}(\text{normalize}(P))$ &=& $P$\\
 $\text{normalize}:\mathbb{R}$ & $\to$ & $\mathbb{R}$\\
 $P$ & $\mapsto$& $\mathbf{P'} = \text{normalize}(P)$ \\
 $\text{denormalize}:\mathbb{R}$ & $\to$ & $\mathbb{R}$\\
 $\mathbf{y'}$ & $\mapsto$& $y = \text{denormalize}(\mathbf{y'})$
\end{tabular}
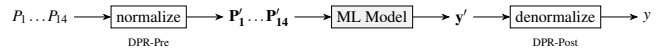
\begin{figure}[htbp]
\centering
\resizebox{\columnwidth}{!}{%
\begin{tikzpicture}[node distance=0.8cm, auto, >=Stealth]
    % Nodes
    \node (raw_x) {$P_1\ldots P_{14}$};
    \node (norm) [draw, rectangle, right=of raw_x] {normalize};
    \node (bold_x) [right=of norm] {$\mathbf{P'_1\ldots P'_{14}}$};
    \node (model) [draw, rectangle, right=of bold_x, fill=gray!10] {ML Model};
    \node (bold_y) [right=of model] {$\mathbf{y'}$};
    \node (denorm) [draw, rectangle, right=of bold_y] {denormalize};
    \node (raw_y) [right=of denorm] {$y$};

    % Arrows
    \draw [->] (raw_x) -- (norm);
    \draw [->] (norm) -- (bold_x);
    \draw [->] (bold_x) -- (model);
    \draw [->] (model) -- (bold_y);
    \draw [->] (bold_y) -- (denorm);
    \draw [->] (denorm) -- (raw_y);
    
    % Labels
    \node [below=0.1cm of norm, font=\scriptsize] {DPR-Pre};
    \node [below=0.1cm of denorm, font=\scriptsize] {DPR-Post};
\end{tikzpicture}
}
\caption{MLC Normalization Pipeline.}
\label{fig:normalization_flow}
\end{figure}

Normalization is a means to ensure that input parameters have the same scale, i.e. have similar training weight.
%In ML frameworks, the normalization uses \emph{calibration data} from the training set to establish $[\min\limits_i(x_i), \max\limits_i(x_i)]$, e.g.
%\begin{tabular}{lll}
%\emph{Calibration data} normalization:&&\\
%$P_1$ range && $P_1$ normalized range\\
%$[-98.2, 49.3]$ & $\mapsto$ & $[0.1, 0.97]$
%\end{tabular}

Instead, we need to generalize to \emph{out-of-sample} data to make DPR compliant with Definition \ref{def:gene}.
We use the lower $\underline{P}$ and upper $\overline{P}$ bounds of $P\in[\underline{P},\overline{P}]$, which are either defined by the input sensor range or the output range specified by the system providing us this input, e.g.
\begin{tabular}{lll}
\emph{Bounds} normalization:&&\\
$P_1$ bounds && normalized $P_1$ bounds\\
$[-100, 50]$ & $\mapsto$ & $[0,1]$\\
\end{tabular}

\textbf{MLC-DPR-Pre}:
The MLC shall normalize $P_i$ input values $\forall i \in [1,14]$:
{\small
\[
\mathbf{P'_i} = \frac{P_i - \underline{P_i}}{\overline{P_i} - \underline{P_i}}
\]
}%
Respectively for the ML model output, we shall specify the bounds of the weight prediction, as the output range which will be used as an input by the user system, e.g.
\begin{tabular}{lll}
weight(kg) bounds && normalized weight\\
$[\text{OEW}, \text{MTOW}]$ & $\mapsto$ & $[0,1]$
\end{tabular}

\textbf{MLC-DPR-Post}:
The MLC shall denormalize $\mathbf{y'}$ output values:
{\small
\[
y = (\overline{y} - \underline{y}) \mathbf{y'} + \underline{y}
\]
}%
{\small
\[
y = (\text{MTOW} - \text{OEW}) \mathbf{y'} + \text{OEW}
\]
}%

Beside MLCR and DPR, the last element coming from the training process is the MLMD.

\section{4. Machine Learning Model Description}
\label{sec:MLMD}
The MLMD describes the model architecture and semantics of the \weight, such that it can be implemented in the 2nd-V, and verified on the target system.
Recurrent Neural Networks (RNN) are usually adequate to this kind of data.
In particular, LSTM based models have demonstrated their efficiency for speech recognition, language translation and other time series.

The MLMD is \emph{built} as an output of the 1st-V \emph{Trained Model} in figure \ref{fig:do178}.
The \emph{build} method is a training framework export to a neutral form, suitable for inference, e.g. Open Neural Network Exchange (ONNX) \cite{ONNX}.
This form contains the Data Flow Graph (DFG) structure, composed of vertices transforming input edges data to output edge data.
The exported ONNX model also contains the values of the Trained Model parameters.

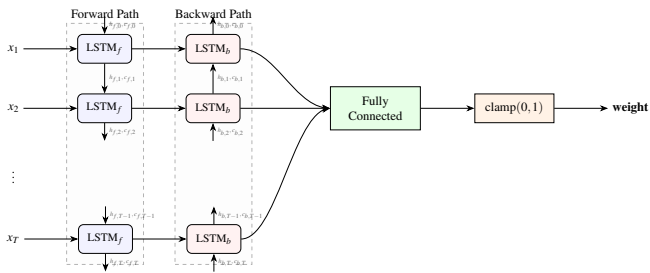
\begin{figure}[htbp]
    \resizebox{1\linewidth}{!}{%
    \centering
    \begin{tikzpicture}[
        node distance=1.2cm and 1.5cm,
        cell/.style={rectangle, draw, rounded corners, minimum width=1.5cm, minimum height=0.8cm, font=\small},
        f_cell/.style={cell, fill=blue!5},
        b_cell/.style={cell, fill=red!5},
        layer_box/.style={draw, dashed, inner sep=0.3cm, fill=gray!5, opacity=0.3},
        arrow/.style={-Stealth, thick},
        state_label/.style={font=\tiny, color=gray!80!black},
        activation/.style={rectangle, draw, fill=orange!10, minimum width=2.2cm, minimum height=0.8cm},
        fc/.style={rectangle, draw, fill=green!10, minimum width=2.5cm, minimum height=1.2cm, align=center}
    ]

    % --- Inputs (Left Column) ---
    \node (x1) {$x_1$};
    \node (x2) [below=of x1] {$x_2$};
    \node (dots_x) [below=of x2] {$\vdots$};
    \node (xT) [below=of dots_x] {$x_T$};

    % --- Forward LSTM Cells (Middle-Left Column) ---
    \node (f1) [f_cell, right=of x1] {$\text{LSTM}_f$};
    \node (f2) [f_cell, right=of x2] {$\text{LSTM}_f$};
    \node (fT) [f_cell, right=of xT] {$\text{LSTM}_f$};
    
    % --- Backward LSTM Cells (Middle-Right Column) ---
    \node (b1) [b_cell, right=of f1] {$\text{LSTM}_b$};
    \node (b2) [b_cell, right=of f2] {$\text{LSTM}_b$};
    \node (bT) [b_cell, right=of fT] {$\text{LSTM}_b$};

    % Connections from Inputs to Cells
    \foreach \i in {1,2,T} {
        \draw [arrow] (x\i) -- (f\i);
        \draw [arrow] (f\i) -- (b\i); % Visual path through forward to backward
    }

    % Forward State Transitions (Downward)
    \draw [arrow] ($(f1.north)+(0,0.5)$) -- node[right, state_label]{$h_{f,0}, c_{f,0}$} (f1.north);
    \draw [arrow] (f1.south) -- node[right, state_label]{$h_{f,1}, c_{f,1}$} (f2.north);
    \draw [arrow] (f2.south) -- node[right, state_label]{$h_{f,2}, c_{f,2}$} ($(f2.south)-(0,0.5)$);
    \draw [arrow] ($(fT.north)+(0,0.5)$) -- node[right, state_label]{$h_{f,T-1}, c_{f,T-1}$} (fT.north);
    \draw [arrow] (fT.south) -- node[right, state_label]{$h_{f,T}, c_{f,T}$} ($(fT.south)-(0,0.5)$);

    % Backward State Transitions (Upward)
    \draw [arrow] ($(bT.south)-(0,0.5)$) -- node[right, state_label]{$h_{b,T}, c_{b,T}$} (bT.south);
    \draw [arrow] (bT.north) -- node[right, state_label]{$h_{b,T-1}, c_{b,T-1}$} ($(bT.north)+(0,0.5)$);
    \draw [arrow] ($(b2.south)-(0,0.5)$) -- node[right, state_label]{$h_{b,2}, c_{b,2}$} (b2.south);
    \draw [arrow] (b2.north) -- node[right, state_label]{$h_{b,1}, c_{b,1}$} (b1.south);
    \draw [arrow] (b1.north) -- node[right, state_label]{$h_{b,0}, c_{b,0}$} ($(b1.north)+(0,0.5)$);

    % --- Output Processing (Right Column) ---
    \node (fc) [fc, right=2.5cm of b2] {Fully\\Connected};
    \node (clamp) [activation, right=of fc] {$\text{clamp}(0,1)$};
    \node (weight) [right=of clamp, font=\bfseries] {weight};

    % Connections to FC
    % We use controls to sweep from the vertical LSTM column to the FC node
    \draw [arrow] (b1.east) .. controls +(1,0) and +(-1,0) .. (fc.west);
    \draw [arrow] (b2.east) -- (fc.west);
    \draw [arrow] (bT.east) .. controls +(1,0) and +(-1,0) .. (fc.west);
    
    \draw [arrow] (fc) -- (clamp);
    \draw [arrow] (clamp) -- (weight);

    % Layer Grouping
    \begin{scope}[on background layer]
        \node [layer_box, fit=(f1) (fT), label=above:Forward Path] {};
        \node [layer_box, fit=(b1) (bT), label=above:Backward Path] {};
    \end{scope}

    \end{tikzpicture}
    }
   \caption{Helicopter weight  Machine Learning Model Description as a dataflow graph including a bidirectionnal LSTM, a fully-connected, and a clamp layer.}
   \label{fig:weight_model}
\end{figure}

The ML Model architecture is described in Figure \ref{fig:weight_model}.
It is a sequence of a bidirectional LSTM cell (forward and backward path), a fully connected (also called linear) layer, and a \texttt{clamp} activation function to limit output  to its specified range.

\begin{comment}
\begin{figure}[!h]
		   \centering
			   \includegraphics[scale=0.5]{weight_model.pdf}
			   \caption{Helicopter weight Machine Learning Model Description, i.e. the dataflow graph of the sequence of operators.}
			   \label{fig:weight_model}
\end{figure}
\end{comment}

The core building block of LSTM RNN is the vanilla LSTM cell \cite{lstm}, defined by the equations \ref{eq:lstm}, where $\{W_f, W_i, W_o, W_c, R_f, R_i, R_o, R_c, B_f, B_i, B_o, B_c\}$ are the model parameters, $\sigma$ the sigmoid gate activation function, \texttt{ReLU} the Rectifier Linear Unit recurrent activation function, $x_t = \{P_1,\ldots,P_{14}\}$ the input parameter vector at time $t$, $h_t, h_{t-1}$ the hidden value at time $t$ and $t-1$  and  $\odot$ the Hadamard product, i.e. element-wise multiplication.
{\small
\begin{align}
	\texttt{ReLU}(x) &= \max(0,x)\nonumber\\
    f_t &= \sigma(W_f \times x_t + R_f \times h_{t-1} + B_f) \nonumber\\
    i_t &= \sigma(W_i \times x_t + R_i \times h_{t-1} + B_i) \nonumber\\
    o_t &= \sigma(W_o \times x_t + R_o \times h_{t-1} + B_o) \nonumber\\
    \tilde{c}_t &= \texttt{ReLU}(W_c \times x_t + R_c \times h_{t-1} + B_c) \nonumber\\
    c_t &= f_t \odot c_{t-1} + i_t \odot \tilde{c}_t \nonumber\\
    h_t &= o_t \odot \texttt{ReLU}(c_t)
\label{eq:lstm}\end{align}
}%
The LSTM cell is recurrent in time, i.e. computed output at time $t$ uses previous cell states $(c_{t-1}, h_{t-1})$.
In our context, the model is used only on a flight event (takeoff), and does not require full flight history.
Before each inference, the LSTM cell state is initialized with $h_0=0$ and $c_0=0$. The model output depends only on 1 sample {}, leading to a deterministic outcome, with the sole knowledge of the current sample.

In the next section, we allocate the MLMD to a single MLM Item Description (MLMID).

The replication criteria are also defined in the MLMD. They specify the level of acceptable discrepancies between the \emph{Trained Model} implementation and the \emph{Deployed Model} implementation. This shall bring confidence that even with those discrepancies, the requirements are still met on the \emph{ODD population}, i.e. comply with the Definition \ref{def:gene}.

\textbf{MLMD-replication-criteria}:
We require a bit-accurate replication: considering the ML model as a black box, the \emph{Deployed Model} output shall be identical to the \emph{Trained Model} output.

\section{5. Machine Learning Model Item Description}
We propose to specify MLMID by \textbf{MLMID-Algorithm}, which describes the ML model algorithm semantics, and \textbf{MLMID-PDI}, which defines the model parameter values. There are two rationales for this choice: - there is a good chance that change occurs less often for algorithms than for model parameters, - the algorithm implementation is designed with source code, and the model parameters are pure data and might be treated as Parameter Data Items (PDI) as per \dosw:
\begin{definition}[Parameter Data Items]
a dataset that influences the behavior of the software (e.g. used as computational data) without modifying the executable object code and is managed as a separate configuration item.
\label{def:pdi}
\end{definition}
\begin{objective}The executable object code is robust with respect to PDI structure and attributes.
\label{obj:PDI-struct}
\end{objective}
\begin{objective}The structure of the LCD allows the PDI to be managed separately.
\label{obj:PDI-separate}
\end{objective}

\textbf{MLMID-Algorithm} The MLMID specifies the ML model semantics, i.e. the algorithm to be implemented, and the interface to ML model parameters PDI to comply with \dosw Objective \ref{obj:PDI-struct}.

In our context, the vanilla LSTM lacks runtime efficiency. The use of $\sigma=\frac{1}{1+e^{-x}}$ gate activation function requires exponential math functions which are computational intensive.

\begin{figure}[!h]
		   \centering
			   \includegraphics[scale=0.6]{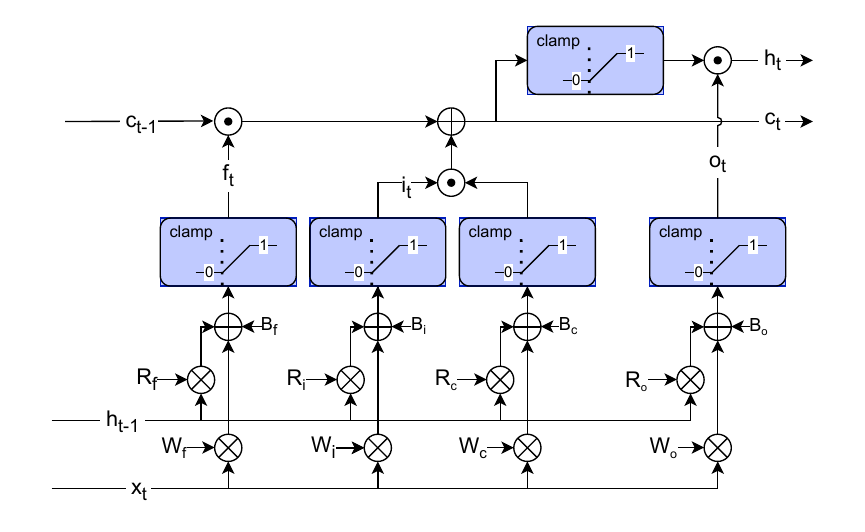}
			   \caption{our LSTM cell informal design}
			   \label{fig:lstm}
\end{figure}

Moreover, the \textbf{MLMID-Algorithm} shall comply with the upward requirement \textbf{MLCR-Robustness}. Unfortunately, the use of \texttt{ReLU} for recurrent activation in the MLMD leads to unbounded data ranges.
Our model is designed with a modified LSTM cell Figure \ref{fig:lstm}, and equation \ref{eq:newlstm}, which replaces \texttt{sigmoid} gates and \texttt{ReLU} activations by \texttt{clamp} gates and activations. Our modified LSTM cell is defined by equation \ref{eq:newlstm}.
{\small
\begin{align}
    \texttt{clamp}(a,b,x) &= \max(\min(b,x),a) \nonumber\\
    f_t &= \texttt{clamp}(0,1,W_f \times x_t + R_f \times h_{t-1} + B_f) \nonumber\\
    i_t &= \texttt{clamp}(0,1,W_i \times x_t + R_i \times h_{t-1} + B_i) \nonumber\\
    o_t &= \texttt{clamp}(0,1,W_o \times x_t + R_o \times h_{t-1} + B_o) \nonumber\\
    \tilde{c}_t &= \texttt{clamp}(0,1,W_c \times x_t + R_c \times h_{t-1} + B_c) \nonumber\\
    c_t &= f_t \odot c_{t-1} + i_t \odot \tilde{c}_t \nonumber\\
    h_t &= o_t \odot \texttt{clamp}(0,1,c_t)
\label{eq:newlstm}\end{align}
}%

As the model semantic is modified by our \texttt{clamp} compared to the MLMD, the accuracy metric is impacted and the \textbf{MLCR-Performance} requirement might not be met anymore.
In this case, we can fine-tune the model, i.e. continue the training to fit the requirement.

Also included in HLR, we specify the Parameter Data Items (PDI).

\textbf{MLMID-PDI}:
We propose to allocate ML model trained parameters to PDI (cf. Definition \ref{def:pdi}).

ML community State Of The Art (SOTA) proposes the use of intermediate representation like ONNX protobuf, LiteRT (TFLite) or Executorch flatbuffer. The export to/from these formats introduces complexity, which increases the risk of design error. Moreover, these formats are not standardized, and are not suitable for long term in-service support.
Instead, we organize the PDI structure as a sequence of multi-dimensional arrays whose dimensions are specified in Interface Document (ID). These arrays are the ML model parameter values $\{W_f, W_i, W_o, W_c,$ $R_f, R_i, R_o, R_c,$  $B_f, B_i, B_o, B_c\}$ for the LSTM layer, and $\{W_{fc}, b_{fc}\}$ for the fully connected layer,  which are the result of the training process and extracted from the ONNX MLMD.
%They are integrated as \dosw Parameter Data Item (PDI) as illustrated on Figure \ref{fig:do178}.

The Interface Document also contains the specification of the ranges of these ML model parameters, such that the ML model algorithm can be robust within those ranges and comply with Objective \ref{obj:PDI-struct}.
For the ML model operations we consider (LSTM and fully-connected), ML model algorithm output is bounded by the norm of the model parameters,i.e. $\|W\|_1, \|R\|_1$.

For this reason, we require that each model parameter $W, R$ satisfy the constraint $\|W\|_1 < \|W_{init}\|_1, \|R\|_1 < \|R_{init}\|_1$, where $W_{init}, R_{init}$ denote the initial parameter values prior to training.
The choice of the value $\|W_{init}\|_1$ as a bound is justified by the observation that training typically oscillates around the initial weights; furthermore, when standard regularization is applied, the parameter norm tends to decrease. Consequently, $\|W_{init}\|_1, \|R_{init}\|_1$ serve as a natural envelope that constrains the weights throughout the training and retraining phases.

The counterpart is that the algorithm specified by \textbf{MLMID-Algorithm} shall be defined, robust and numerically stable for any ML model parameter inside the envelope $\|W\|_1 < \|W_{init}\|_1$.

Once all MLC elements are ready for the software design, we are able to define the software architecture including the MLC logical architecture.

\section{6. ML Software Item Architecture}
  \begin{figure}[h!]
  \centering
    \resizebox{1\linewidth}{!}{%
  \includegraphics{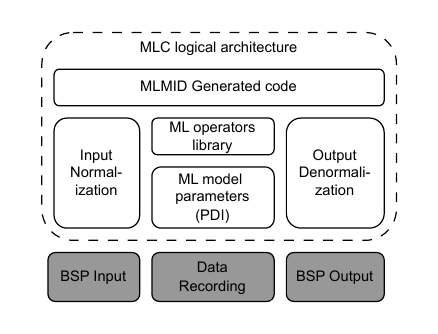}
  }
	\caption{Software item architecture\label{fig:SWitem}
}
\end{figure}

The software architecture in Figure \ref{fig:SWitem} breaks down the software item into software sub-components which are allocated to the HLR:
\begin{itemize}\setlength{\itemsep}{0pt}
\item the MLC logical architecture composed of:
\begin{itemize}[before*={\mbox{}\vspace{-.75\baselineskip}}]
\setlength{\itemsep}{0pt}
\item the generated code implementing the MLMID dataflow,
\item the ML operators or kernels implementing ML algorithm,
\item the input normalization and output denormalization, specified by DPR
\item the ML model trained parameters as \dosw Parameter Data Item (PDI)
\end{itemize}
\item the Board Support Package (BSP) to connect system input output (I/O),
\item the data recording feature for in-service monitoring.
\end{itemize}

Having defined the HLR and the MLC architecture,  the next step is to consider the Low Level Requirements (LLR) as per Figure \ref{fig:do178}.
LLR are \dosw LCD, which specify \emph{how} the software implements HLR:
\begin{itemize}\setlength{\itemsep}{0pt}
\item how the model is computed, i.e. the detailed design of the MLMID operations specified by upward requirement \textbf{MLMID-Algorithm}.
\item model parameter values (weight and bias) specified by  \textbf{MLMID-PDI} and their integrity checksum.
\item refinements HLR corresponding to DPR (\textbf{MLC-DPR-Pre}, \textbf{MLC-DPR-Post}).
\item refinements HLR corresponding to BSP I/O and recording requirements. These are out of this paper scope.
\end{itemize}

%\todo[inline]{BL: C'est la troisième représentation du LSTM, et la traceability entre les trois est pas évidente. Est-ce que tu peux réorganiser les boîtes Scade pour correspondre au flot de la Figure 8?
%NV: TODO mais celà montre qu'il y a bien un étage de design, que ce n'est pas du 1:1}

\begin{figure}[!h]
		   \centering
			   \includegraphics[scale=0.9]{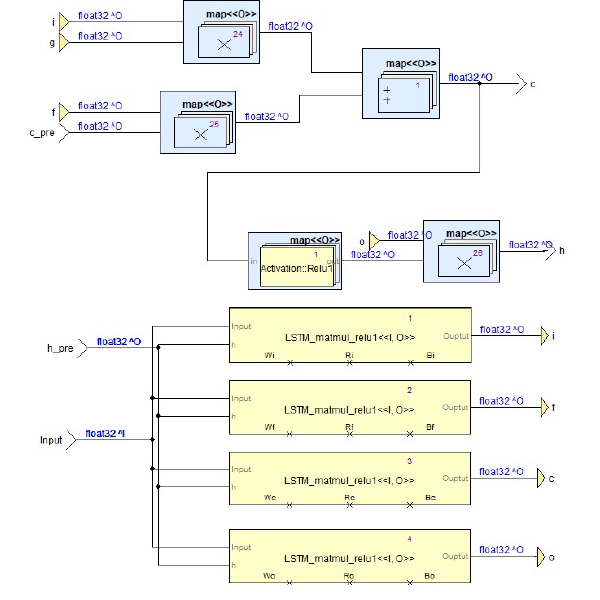}
			   \caption{\weight model LSTM cell implementation with Scade}
			   \label{fig:scade}
\end{figure}

To implement \textbf{MLC-DPR-*} and \textbf{MLMID-Algorithm}, we choose to formalize LLR with the ANSYS Scade tool as in \cite{DASC2025valot}, and use their \dotool code generator.
The Figure \ref{fig:scade} presents the formal description of the LSTM cell for floating point 32 bits data. On top the Hadamard products are performed element-wise on O-size vectors using the \texttt{map}$\ll$\texttt{O}$\gg$ operator, then the \texttt{clamp(0,1)} namely \texttt{Activation:Relu1}.
At the bottom of the Figure, the matrix multiplications with the PDI ML model parameters $\{W_f, W_i, W_o, W_c,$ $R_f, R_i, R_o, R_c,$  $B_f, B_i, B_o, B_c\}$, defined as Scade \emph{imported constants}, which complies to Objective \ref{obj:PDI-separate}, and resolved at link time.

Following \textbf{MLMID-Algorithm} code generation, the source code is compiled for the target processor and linked into the final software executable alongside other software components.
The Trained Model parameters \textbf{MLMID-PDI} are exported to a binary file for the target.
In this scenario, two approaches are considered: i) we may use a bit-accurate literal representation, such as the hexadecimal floating-point format supported by C99 and IEEE-754 (e.g., \texttt{0x1.0e4978p+2}).
ii) constants can be exported to a raw binary file and converted into an Executable and Linkable Format (ELF) object using the GNU \texttt{objcopy} tool. While the former is endianness-agnostic, the latter significantly reduces compilation time; however, it may necessitate an endianness swap if the target architecture is big-endian, as training environments are typically little-endian.

In addition to these methods, we compute an integrity checksum before the binary conversion. This checksum is verified in the target before model inference as a PowerOn Self Test (POST).

Once MLC implementation has led to an \emph{executable},
we can process its \emph{verification} in Figure \ref{fig:do178}.
The implementation shall be verified to demonstrate its compliance with its requirements (MLCR) and to the \arp and \dosw objectives.

\textbf{Low Level Test (LLT)}
The \textbf{MLMID-Algorithm} and \textbf{MLC-DPR-*} are implemented by formal Scade modeling and the generated code is qualified and is therefore not verified by LLT.
Other LLR related to libraries, BSP, data recording and are verified by LLT; these are out of this paper scope.

\textbf{Data Flow and Control Flow analysis}
The code generator automatizes the data flow and control flow, and we take advantage of \dotool qualification, to remove the \dosw control flow and data flow verification activity.

\section{7. High Level Tests}
High-Level Tests (HLTs) represent the most critical artifacts of the software item. 
They are necessary to demonstrate the compliance of the implementation with the HLR, i.e. DPR and MLCR.

We first define experimental target configurations for HLT setup.
To assess HLTs, we implement the \emph{Deployed Model} for 2 possible target hardware:
\begin{itemize}\setlength{\itemsep}{0pt}
\item
NXP QorIQ\textsuperscript{\textregistered}  T1042 CPU e5500 core \cite{e5500}, 900MHz, Asterios RTOS \cite{asterios}.
For this target, we evaluate 2 compiler configurations: i) Windriver Diab with conservative optimization options, with optional \texttt{-Xfp-pedantic}, ii) CompCERT \texttt{-Os} optimize for size or \texttt{-O3} optimize for speed.
\item
TMS570LC43 ARM\textsuperscript{\textregistered}  Cortex\textsuperscript{\textregistered}  R5F \cite{cortex_r5} 300MHz, with the TI ARM compiler, \texttt{-O2 VFPv3D16}, with optional ARM-thumb 16 bits instructions.
\end{itemize}

We propose a number of experimental configurations of the \emph{Deployed Model}, each denoted as follows:
    \usecase{tar}{comp}{repr}{optim}.
\begin{itemize}\setlength{\itemsep}{0pt}
    \item \ucf{tar} identifies the considered target hardware \{\texttt{T1042}: NXP T1042 cpu, or \texttt{TMS570}: Texas Instrument Hercules\}.
    \item \ucf{comp} identifies the compiler  {\texttt{ccomp}: \texttt{CompCERT}, or \texttt{diab}: Windriver Diab}
    \item \ucf{repr} identifies the machine representation \{\texttt{fp32}: floating point 32 bits, \texttt{fp32x2}: 2-vector fp32, \texttt{I16}: integer 16 bits, \texttt{bf16}: the bfloat16 format,.i.e. the truncated \texttt{fp32}\}.
    \item \ucf{optim} identifies the compiler optimization option \{\texttt{Os} : optimized for size, \texttt{O3}: optimized for speed, \texttt{pedantic}: strict IEEE754 (disables FMA)\}.
	\item \ucf{ds} identifies the dataset \{\texttt{lhs}: latin hypercube sampling of input space, \texttt{train}: the training set, \texttt{test}: the test set\}.
	\item \ucf{repr} identifies the machine representation \{\texttt{fp32}: floating point 32 bits, \texttt{bf16}: bfloat16 is a truncated mantissa of \texttt{fp32}, \texttt{i16}: integer 16 bits\}.
\end{itemize}

We verify the first requirement linked to a design constraint which limits the CPU resource allocated to the ML Model.

\textbf{HLT-MLM-Execution-time}:
the software implementation is made of statically bounded loops to compute matrix multiplication, and some balanced branches for \texttt{clamp} decisions.
Therefore, there is no source of execution time variability at the software level of the MLC.
Nevertheless, the software item also deals with other requirements, which can include some variabilities.
In this paper, we will focus on the ML Model contribution, and specifically to its execution time measurement on the target.
In a full development plan for \dosw, we would address Worst Case Execution Time computation, which would account for additional margins (e.g. cache effects, multicore interference channels).

The ML Model inference execution time is gathered for both target configurations.

\begin{figure}[!h]
		   \centering
%		   \begin{subfigure}{0.49\linewidth}
			   \includegraphics[scale=0.5]{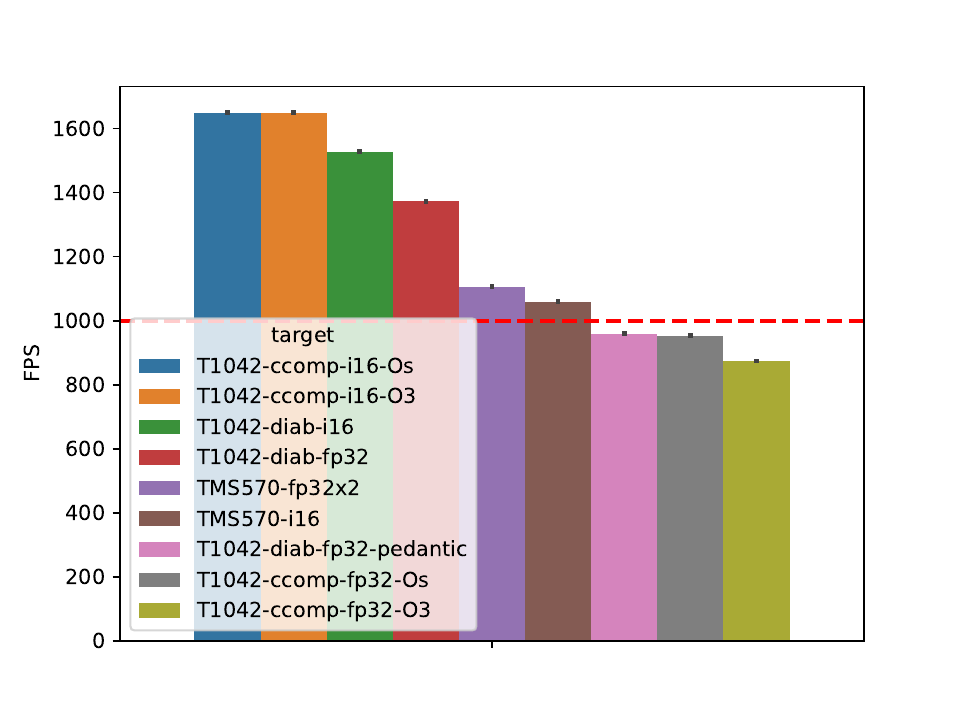}
%			   \vspace{-1em}
			   \caption{ML Model inference per second (higher is better), \textbf{MLCR-MLM-Execution-time} requires 1 ms, i.e. 1000 FPS. \label{fig:fps}}
%		   \end{subfigure}
%		   \hfill
\end{figure}

The experimental measurement of running time is reported in Figure \ref{fig:fps}. The Frame Per Seconds (FPS) are computed as $\text{FPS}=\frac{1}{t_1 - t_0}$, where $t_0, t_1$ are respectively the timestamps of the model \emph{weight} function begin and end (excluding normalization and denormalization).

\textbf{Analysis of the execution time measurements.}
Thanks to a higher operating CPU frequency, the best performance is obtained by \texttt{T1042} target with integer 16 bits (\texttt{I16}) implementation compiled with \texttt{CompCERT} compiler, with optimizations.
The \texttt{I16} variant uses 2x less memory than \texttt{fp32}.
Despite a lower operating frequency, the \texttt{TMS570} CPU owns a Single Instruction Multiple Data (SIMD) capability, which handles \texttt{fp32} as a 2-dim vector (\texttt{fp32x2}) and doubles the computing bandwidth. On the counterpart, it uses twice the memory bandwidth as the \texttt{I16}.
When the model parameters fit into the data cache of the CPU, the memory bandwidth is not the bottleneck, then the \texttt{fp32} is the best tradeoff.
When model parameters do not fit into data cache, low precision integers (\texttt{I16}) provide the best FPS, but the complexity to implement and demonstrate correct integer parameter scaling might not be worth it. In our context, real-time constraint is not at stake.

The software item's WCET is computed and verified among the software item time budget. This is out of this papers' scope.

%\todo[inline]{BL: Il manque des \texttt{optim} dans certaines configurations de la Figure~\ref{fig:fps}
%NV: oui, TODO}
\textbf{HLT-MLC-DPR-Pre} and \textbf{HLT-MLC-DPR-Post} are test cases in legacy of \dosw requirement based testing, i.e. some nominal cases in the range of $P_1..P_{14}$, some cases at the range boundaries, and some cases out of range (robustness cases).

Since MLC Requirements are typically based on statistical metrics, establishing trust in the software requires more than a simple \emph{passed} status. Our confidence is equally rooted in the methodologies used within the HLTs to verify these requirements and the rigor of the resulting analysis.

\textbf{HLT-MLCR-Performance}:
The computed metrics using the \emph{Test dataset} and the \emph{Trained Model} are:
\begin{center}
\begin{tabular}{lll}
\hline
Metric&computed value& requirement threshold\\
\hline
$\frac{\text{MAE}}{\text{MTOW}}$ & $=1.9\cdot 10^{-2}$ & $<3\cdot 10^{-2}$\\
\hline
$\text{R}^2$ & $=84.9\%$ & $> 80\%$\\
\hline
\end{tabular}
\end{center}
Therefore our implementation complies with the to \textbf{MLCR-Performance} requirements for the \emph{Trained Model}.
We take credit for this assessment for the \emph{Deployed Model} thanks to our \textbf{MLMD-replication-criteria}.

\textbf{HLT-oos-confidence-interval}:
we propose to tackle the specified \emph{generalization} definition \ref{def:gene} with a Confidence Interval (CI) of the MAE and $\text{R}^2$ with probability of 99\% using the \emph{bootstrapping} method \cite{bootstrap_ci}, using 10000 resampling of the random variable $|y_i - \hat y_i|$ which is used to compute MAE and $\text{R}^2$ metrics in Equations \ref{eq:MAE} and \ref{eq:R2}.

\begin{center}
\begin{tabular}{lll}
\hline
Metric&computed value& requirement threshold\\
\hline
$\frac{\text{MAE}}{\text{MTOW}}$ & $< 2\cdot 10^{-2}$ & $< 3.15 \cdot 10^{-2}$\\
\hline
$\text{R}^2$ & $> 82.8\%$ & $> 76\%$\\
\hline
\end{tabular}
\end{center}
\emph{Note: we are specifically interested in the CI upper bound for MAE and lower bound for $\text{R}^2$.}

We verified \textbf{MLCR-oos-confidence-interval} with the \emph{bootstrapping} approach. Being more computing intensive than the \emph{frequentist} approach, it does not assume an underlying Normal distribution, and is more robust to outliers.

Another approach called \emph{frequentist}, requires a Normal distribution of $|\hat y - y|$.
It establishes a relationship between the number of samples $n$ of the \emph{Test dataset}, and the distribution attributes (mean and variance). The confidence level is defining the $z$ parameter of the normal distribution $N(0,1)$: probability $P(-Z<z<Z)<99\%$.
\[
n>\left(\frac{z\cdot\sigma}{5\%\cdot \text{MAE}}\right)^2
\]
with $z = P^{-1}(99.5\%)=2.576$ and $\sigma$ the standard deviation of the \emph{ODD population}. $\sigma$ being unknown, we choose to approximate it with $\sigma(|\hat y - y|)$, the standard deviation of the ML model prediction error.
\[
n>\left(\frac{2.576\cdot\sigma(|\hat y - y|)}{5\% \cdot \text{MAE}} \right)^2
\]
We measured on the \emph{Test dataset} $\sigma(|\hat y - y|) = 0.05$
\begin{equation}
\small
n >1811
\label{eq:mindataset}
\end{equation}
Our \emph{Test dataset} contains $n=2110$ samples which satisfies \ref{eq:mindataset}.

\textbf{HLT-MLCR-Stability}:
the High Level Test of MLC stability verifies the requirement \textbf{MLCR-Stability} in a number of \emph{experimental configurations}, each denoted as follows:
   \ucf{ds}-\ucf{repr}.

\begin{figure}[!h]
		   \centering
			   \includegraphics[scale=0.25]{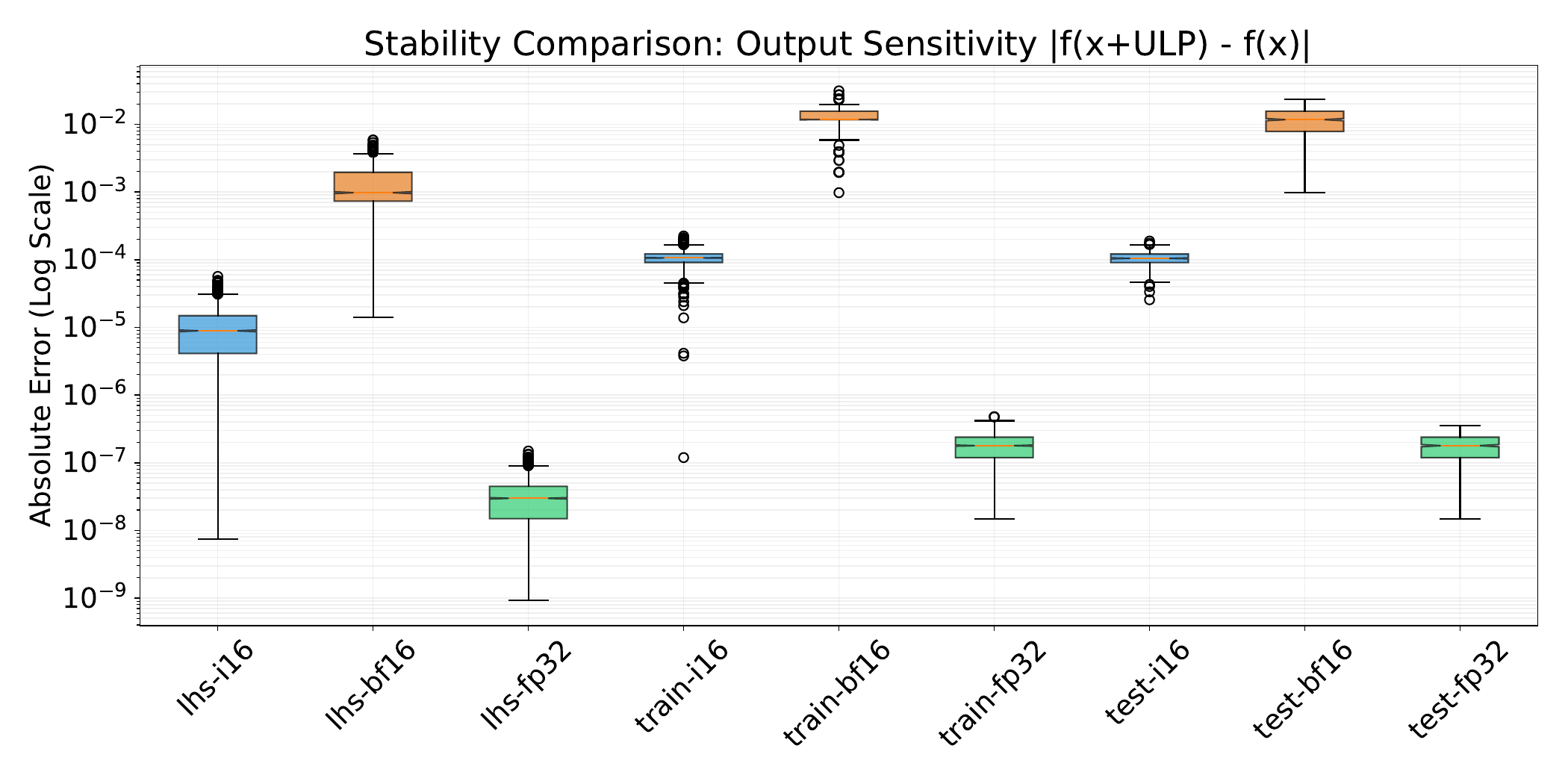}
			   \caption{HLT-MLCR-Stability: distribution of the stability error $|\text{weight}(x+\epsilon) - \text{weight}(x)|$\label{fig:stability}}
\end{figure}

The results are plotted as a distribution in Figure \ref{fig:stability}. Unsurprisingly, the \texttt{i16} is more stable than \texttt{bf16} because it allocates more bits to accuracy.

The stability shall also satisfy the \emph{generalization} definition \ref{def:gene}.
We apply the bootstrapping method \cite{bootstrap_ci} to compute the upper bound of the Confidence Interval at 99\%.
We assessed that both \texttt{i16} and \texttt{fp32} comply with the \textbf{MLCR-Stability} requirement.

\textbf{HLT-MLCR-Robustness}:
we make use of formal methods, namely Interval Bound Propagation (IBP) \cite{IBP} and the autoLiRPA tool \cite{xu2020automatic} to automate the ML model intermediate layer ranges computation, and demonstrate that they fit the chosen implementation data types.
This is the pillar which ensures that the implementation of \textbf{MLMID-Algorithm} is robust to the PDI value's range.

\textbf{HLT-MLMD-replication-criteria}:
we summarize in a table the status of the replication criteria.
For any input $x'_i$ in the \emph{Test dataset}, bit-accurate is \textbf{Y} if $\text{weight}_{tm}(x'_i) = \text{weight}_{dm}(x'_i)$, $tm$ denotes the \emph{Trained Model}, and $dm$ the \emph{Deployed Model}.

\begin{center}
{\small
\begin{tabular}{ll}
\hline
Configuration&bit-accurate\\
\hline
\texttt{T1042-ccomp-I16-Os}&\textbf{Y}\\
\texttt{T1042-ccomp-I16-O3}&\textbf{Y}\\
\texttt{T1042-diab-I16}&\textbf{Y}\\
\texttt{T1042-diab-fp32}&N\\
\texttt{T1042-diab-fp32-pedantic}&\textbf{Y}\\
\texttt{TMS570-fp32x2}&\textbf{Y}\\
\texttt{TMS570-I16}&\textbf{Y}\\
\texttt{T1042-ccomp-fp32-Os}&\textbf{Y}\\
\texttt{T1042-ccomp-fp32-O3}&\textbf{Y}\\
\hline
\end{tabular}
}%
\end{center}

The \texttt{T1042-diab-fp32} configuration does not satisfy this requirement because it uses Fuse Multiply Add (FMA) instructions leading to specific \emph{Target model} rounding effects which prevents a bit-accurate replication of the \emph{Trained Model}.

\textbf{HLT-MLMID-Algorithm}:
the algorithm implemented by the Scade model is reviewed (inspection) against the ONNX MLMD.

\textbf{HLT-MLMID-PDI}:
the MLMID parameters integrity is computed on the binary parameter file and compared to the reference checksum.
The $\|W\|_1< \|W_{init}\|_1$ is statically verified for any of the ML model parameters (i.e. LSTM, and Fully-connected layers).

\textbf{MLC Integration and Verification}:
in our MLC, there is a single ML Model. There is no integration at this level.
The MLC verification takes credit for the software item HLT verification.

\section{Related work}
The \weight use case described in this paper, is also addressed in \cite{weightestim, weight,VFS2025ammar, abraham2009flight,iele2016feasibility}.
We provide some insights to design assurance development of such use cases, with the objective to comply with \arp, and \dosw.
For this work, we choose as a start point paper \cite{weightestim}.

The formalization of the MLMD was addressed by \cite{FormalDescMLmodel,sonnx,DASC2025valot}.
\cite{FormalDescMLmodel} did not explore how to export a MLMD, and how to
verify its compliance to \arp objectives.
\cite{DASC2025valot} introduced the specification of requirements based on ML metrics thresholds. We extend their approach with the confidence interval (CI) and the bootstrapping method verification.
\cite{sonnx} introduces a safety profile based on rigorous formal definitions. 
While this approach is promising, its practical utility depends on the availability of qualified tools for importing/exporting SONNX models and the support of qualified code generators. However, the method presents two significant drawbacks: first, it introduces a proprietary description format that lacks broad adoption within the ML community. Second, the rigid formalization of operator algorithms imposes severe constraints on the source code, precluding hardware-specific optimizations and significantly reducing computational efficiency. In our current approach we use standard ONNX to communicate the MLMD from 1st-V to 2nd-V but we formally design the \textbf{MLMID-Algorithm} with Scade modeler and extract ML model parameter values with a python script to build \textbf{MLMID-PDI}.

While \cite{acasxu_erts2026} adopts a similar approach for performance and stability characterization, we provide a more comprehensive treatment of requirement development, specifically regarding the specification of robustness and stability requirements.

The paper \cite{alexis_military} addresses a comparable approach on object detection tasks focusing on the 1st-V of the \arp W-shape development process, while we focus on the 2nd-V of the \arp W-shape.

The Interval Bound Propagation (IBP) \cite{IBP} is usually used for adversarial training or verification. We use it to compute the ranges of intermediate layers of neural networks and assess their compliance with layer bounds requirements.

\section{Conclusion}
We introduced the development life-cycle of a Machine Learning regression model used as a virtual sensor of helicopter weight.
We described step by step the main development activities and artifacts from requirements to implementation and verification on typical avionic target hardware. 
We clarified the \emph{generalization} term and used bootstrapping and frequentist methods to demonstrate it for regression metric requirements.
We proposed a robust ML architecture with the introduction of \emph{clamp} in the LSTM cell, and verified its robustness with a formal method (IBP).
We evaluated several software implementation configurations on avionics hardware compatible with our legacy development tools and standards.
We believe that the presented  artifacts of our ML model are a good start point of the  certification process according to \arp and \dosw objectives.

It also paves the way to design other virtual sensor regressors use cases.
In a future work, where the \emph{Out of distribution detection} model would be developed together with the \weight model, we would integrate them into the same MLC as they share the same normalization pre-processing requirement.
\bibliographystyle{ahs}
\bibliography{main}

% File generated by: AHS Paper Bibliographic Style File
% Version 1.20, 9/16/2023
% Written by Matt Floros
% Last Modified by Matt Floros

\begin{thebibliography}{10}
\newcommand{\enquote}[1]{``#1''}

\bibitem{Easaconcept}
{EASA}, \enquote{{Concept Paper: guidance for Level 1 \& 2 machine learning
  applications - Proposed Issue 02},} , 2024.

\bibitem{ARP6983}
{EUROCAE WG-114/SAE joint group}, \enquote{{ED-324 Process Standard for
  Development and Certification/Approval of Aeronautical Safety-Related
  Products Implementing {AI} },} open consultation, 2025.

\bibitem{weightestim}
Mechouche, A., Valot, N., and Fabre, L., \enquote{Towards Learning Assurance
  for In-Flight Machine Learning-Based Helicopter Weight Estimator,} Vertical
  Flight Society 82th Annual Forum Proceedings, West Palm Beach, FL, 2026.

\bibitem{ONNX}
Bai, J., Lu, F., Zhang, K., \emph{et~al.}, \enquote{{ONNX: Open Neural Network
  Exchange},} \url{https://onnx.ai/}, 2019.

\bibitem{lstm}
Hochreiter, S., and Schmidhuber, J., \enquote{Long Short-Term Memory,}
  \emph{Neural Comput.}, 1997, pp.~1735--1780.

\bibitem{DASC2025valot}
Valot, N., Fabre, L., Lesage, B., Mechouche, A., and Pagetti, C.,
  \enquote{{Implementation of airborne ML models with semantics preservation},}
  {44th Digital Avionics Systems Conference (DASC)}, September 2025.

\bibitem{e5500}
\enquote{{NXP e5500 Core Reference Manual, Rev. 3},} , Nov 2012.

\bibitem{asterios}
Methni, A., Ohayon, E., and Thurieau, F., \enquote{{ASTERIOS Checker : A
  Verification Tool for Certifying Airborne Software},} {10th European Congress
  on Embedded Real Time Systems (ERTS 2020)}, 2020.

\bibitem{cortex_r5}
\enquote{{Texas Instrument Cortex\textsuperscript{\textregistered} -R5 and
  Cortex-R5F Technical Reference Manual},} .

\bibitem{bootstrap_ci}
Efron, B., \enquote{Better bootstrap confidence intervals,} \emph{Journal of
  the American statistical Association}, 1987.

\bibitem{IBP}
Gowal, S., Dvijotham, K., Stanforth, R., Bunel, R., Qin, C., Uesato, J.,
  Arandjelovic, R., Mann, T.~A., and Kohli, P., \enquote{On the Effectiveness
  of Interval Bound Propagation for Training Verifiably Robust Models,}
  \emph{CoRR}, 2018.

\bibitem{xu2020automatic}
Xu, K., Shi, Z., Zhang, H., Wang, Y., Chang, K.-W., Huang, M., Kailkhura, B.,
  Lin, X., and Hsieh, C.-J., \enquote{Automatic perturbation analysis for
  scalable certified robustness and beyond,} \emph{Advances in Neural
  Information Processing Systems}, Vol.~33, 2020.

\bibitem{weight}
Mechouche, A., Rocher, A., and Aubin, V., \enquote{Method for training at least
  one artificial intelligence model for estimating the mass of an aircraft
  during flight based on utilisation data,} US Patent US20240005207A1 App.
  18/209,183, 2024, European Patent EP4300053B1.

\bibitem{VFS2025ammar}
Mechouche, A., Houles, M., Gallimard, C. D.~C., and Maisonneuve, P.-L.,
  \enquote{{On the Trustworthiness of Machine Learning Models in Health and
  Usage Monitoring of In-Service Helicopters},} {81th Vertical Flight Society
  Forum (VFS)}, May 2025.

\bibitem{abraham2009flight}
Abraham, M., and Costello, M., \enquote{In-flight estimation of helicopter
  gross weight and mass center location,} \emph{Journal of Aircraft}, Vol.~46,
  2009.

\bibitem{iele2016feasibility}
Iele, A., Leone, M., Solimeno, R., Grasso, C., Persiano, G., Cutolo, A., and
  Cusano, A., \enquote{A feasibility analysis for the development of novel
  aircraft weight and balance monitoring systems based on fiber Bragg grating
  sensors technology,} Proceedings of the 8th European Workshop on Structural
  Health Monitoring (EWSHM 2016), Bilbao, Spain, 2016.

\bibitem{FormalDescMLmodel}
Gauffriau, A., De~Albuquerque~Silva, I., and Pagetti, C., \enquote{{Formal
  description of ML models for unambiguous implementation},} {12th European
  Congress on Embedded Real Time Software and Systems (ERTS)}, 2024.

\bibitem{sonnx}
Jenn, E., Souyris, J., Belfy, H., Correnson, L., Turki, M., Valot, N., and
  Vedrine, F., \enquote{{SONNX: Towards an ONNX Profile for critical systems},}
  {Proceeding of ERTS 2026 on Embedded Real Time Systems}, 2026.

\bibitem{acasxu_erts2026}
Gabreau, C., Valot, N., Perotto, F., and Pagetti, C., \enquote{{Implementation
  and certification of ML-based surrogate models in avionic systems},}
  {Proceeding of ERTS 2026 on Embedded Real Time Systems}, 2026.

\bibitem{alexis_military}
de~Cacqueray, A., Ribas~de Amaral, J., and Capdevila~Llompart, C.,
  \enquote{{Overview of Initiatives Suitable for Learning Assurance of AI-Based
  Military Products and Identified Challenges based on Use Cases Analysis},}
  Proceeding of Deutsche Gesellschaft für Luft- und Raumfahrt -
  Lilienthal-Oberth e.V., Bonn, 2025.

\end{thebibliography}

\end{document}